\documentclass{llncs}

\usepackage[preprint]{eccv}

\usepackage{eccvabbrv}

\usepackage{graphicx}
\usepackage{booktabs}

\usepackage[accsupp]{axessibility}  
\usepackage[table]{xcolor}

\usepackage{placeins}
\usepackage{booktabs}
\usepackage{siunitx}
\usepackage{hyperref}

\usepackage{orcidlink}

\begin{document}

\title{PVI: Plug-in Visual Injection for Vision-Language-Action Models} 

\titlerunning{PVI}




\author{Zezhou Zhang \and
Songxin Zhang \and
Xiao Xiong \and
Junjie Zhang \and
Zejian Xie \and
Jingyi Xi \and
Zunyao Mao \and
Zan Mao \and
Zhixin Mai \and
Zhuoyang Song$^\dagger$ \and
Jiaxing Zhang$^\dagger$}

\authorrunning{Z.~Zhang et al.}

\institute{Lionrock AI Lab, China Merchants Group, Hong Kong, China\\
\email{zezhou.zhang@alumni.pku.edu.cn}\\
$^\dagger$Corresponding authors: \email{\{songzhuoyang, zhangjiaxing\}@cmhk.com}}

\authorrunning{Z. Zhang et al.}

\maketitle

\begin{abstract}

    VLA architectures that pair a pretrained VLM with a flow-matching action expert have emerged as a strong paradigm for language-conditioned manipulation. Yet the VLM, optimized for semantic abstraction and typically conditioned on static visual observations, tends to attenuate fine-grained geometric cues and often lacks explicit temporal evidence for the action expert. Prior work mitigates this by injecting auxiliary visual features, but existing approaches either focus on static spatial representations or require substantial architectural modifications to accommodate temporal inputs, leaving temporal information underexplored. We propose Plug-in Visual Injection (PVI), a lightweight, encoder-agnostic module that attaches to a pretrained action expert and injects auxiliary visual representations via zero-initialized residual pathways, preserving pretrained behavior with only single-stage fine-tuning. Using PVI, we obtain consistent gains over the base policy and a range of competitive alternative injection strategies, and our controlled study shows that temporal video features (V-JEPA2) outperform strong static image features (DINOv2), with the largest gains on multi-phase tasks requiring state tracking and coordination. Real-robot experiments on long-horizon bimanual cloth folding further demonstrate the practicality of PVI beyond simulation.

\end{abstract}

\section{Introduction}
\label{sec:intro}

Vision-Language-Action (VLA) models offer a scalable pathway toward
generalist robot manipulation by unifying language comprehension with
visuomotor control~\cite{ref1,ref2,ref3,ref4,ref5,ref6}.
A prevailing design pairs a Vision-Language Model (VLM) for semantic
understanding and instruction grounding with a diffusion or
flow-matching Transformer as the action expert, which generates continuous action sequences
conditioned on the VLM's representations~\cite{ref7,ref8,ref9,ref10,ref11,ref12}.
However, this factorization carries a structural limitation: the VLM's
language backbone, optimized for semantic abstraction rather than geometric
fidelity, tends to compress away fine-grained spatial
cues---object geometry,
edge alignments, and grasp affordances---before
they ever reach the action expert~\cite{ref1,ref13,ref14}, precisely
the information that dexterous manipulation demands.

A natural response to this bottleneck is to enrich the visual information
available to the action network beyond what the VLM pathway preserves.
Prior work pursues this in two complementary ways: fusing
geometry-grounded or spatially enriched features into the VLM's visual
tokens before semantic processing~\cite{ref23,ref24},
or bypassing the VLM entirely by injecting auxiliary visual
representations directly into the action expert via side
branches~\cite{ref14,ref15,ref16,ref17,ref33}. Both strategies
recover meaningful geometric detail and have demonstrated improvements in
manipulation precision. Yet both strategies share a common limitation: the injected
representations are often static, capturing the spatial state
of a scene at a single instant while remaining blind to the temporal
dynamics---motion continuity, progressive contact, substep
transitions---that closed-loop manipulation inherently demands.

Recent work has begun to explore richer temporal and video-based
representations within VLA models, demonstrating that motion and
contact dynamics provide complementary information beyond static
features~\cite{ref19,ref25,ref26,ref47}. However, these approaches typically require substantial architectural modifications to the pretrained backbone or the VLM-to-action interface to accommodate temporal inputs, raising a more fundamental question: how can we inject temporally aware visual information in a way that is lightweight, minimally invasive, and compatible with existing pretrained VLA weights?

\begin{figure}[tb]
  \centering
  \includegraphics[width=\linewidth]{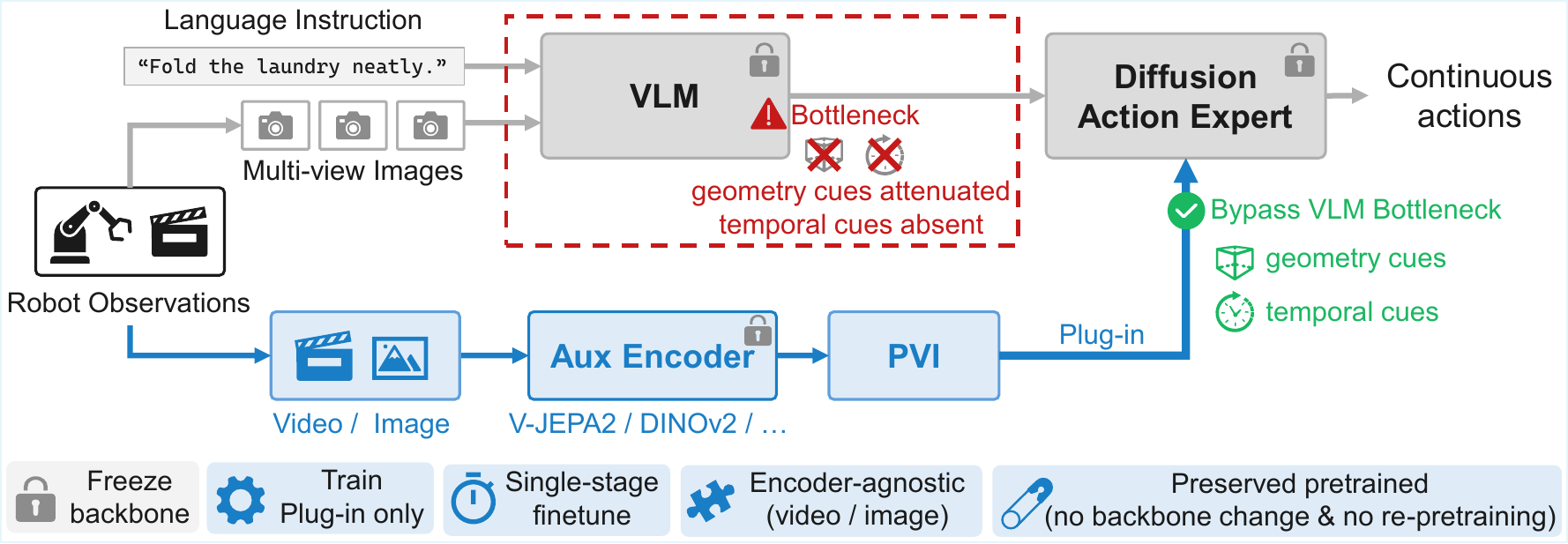}
  \caption{\textbf{Overview of Plug-in Visual Injection (PVI).} Typical VLAs condition the VLM on static images with language, providing limited temporal context; moreover, the VLM’s output representations may under-emphasize fine-grained geometric cues. PVI bypasses this bottleneck by injecting auxiliary visual representations directly into the frozen action expert via a trainable plug-in, with no backbone modification and no re-pretraining required.}
  \label{fig:fig_1}
\end{figure}

We address this challenge with Plug-in Visual Injection (PVI), a
lightweight module that injects auxiliary visual representations directly
into the flow-matching action expert of a pretrained VLA without modifying
its architecture or requiring retraining from scratch, as shown in \cref{fig:fig_1}. PVI introduces a
frozen auxiliary visual encoder alongside a trainable copy branch that
mirrors the action expert's architecture and conditioning interface. The
copy branch is conditioned on projected auxiliary visual features rather
than the VLM representations; its layer-wise hidden states are mapped by
zero-initialized injection layers and added residually to the
corresponding layers of the frozen main action expert, allowing auxiliary
information to influence denoising across the full depth of the policy
while preserving pretrained representations. This design enables stable,
single-stage fine-tuning through the injection pathway alone. Crucially,
the conditioning interface is encoder-agnostic, admitting static,
temporal, or combined visual encoders under a unified injection
mechanism, decoupling the injection infrastructure from any particular
representation choice.

We instantiate PVI on GR00T N1.5\cite{ref10} and evaluate it on simulated bimanual
manipulation tasks spanning diverse primitives, complemented by
real-robot experiments on deformable-object manipulation.
Our contributions are threefold:
\begin{itemize}
  \item \textbf{PVI framework.} We propose Plug-in Visual Injection
  (PVI), a stable, minimally invasive, and encoder-agnostic
  visual-injection framework for pretrained VLAs. Via a dual-pathway
  design with layer-wise zero-initialized residual injection, PVI injects
  auxiliary representations into a frozen action expert while preserving
  its pretrained behavior, requiring only single-stage fine-tuning of the
  injection pathway. Its encoder-agnostic interface admits static,
  temporal, or combined visual encoders under a unified mechanism,
  decoupling the injection infrastructure from any particular
  representation choice.

  \item \textbf{Effective injection via systematic design comparison.}
   We systematically evaluate PVI against a range of plausible injection
  strategies across diverse bimanual simulation tasks. PVI improves the
  average success rate from 35.7\% to 59.7\% (+24.0 percentage points),
  outperforming competitive alternatives. Real-robot experiments further validate the effectiveness of PVI beyond simulation, where PVI successfully drives long-horizon bimanual cloth folding that demands precise two-arm coordination over a highly deformable object.

  \item \textbf{Representation study.} Leveraging PVI as a controlled
  testbed, we systematically compare temporal video encoders, static
  image encoders, and their combinations. Temporal video features
  (V-JEPA2\cite{ref44}) consistently yield larger gains than strong static baselines
  (DINOv2\cite{oquab2023dinov2}), especially on tasks requiring state tracking and multi-phase
  coordination, providing direct empirical evidence that temporal dynamics
  carry complementary information beyond what static representations can
  supply.
\end{itemize}

\section{Related Works}
\subsection{VLA Models with Diffusion/Flow-Matching Action Experts}

Early VLAs cast action generation as autoregressive token prediction,
inheriting the semantic generalization of large VLMs but relying on
action discretization, which can limit fine-grained continuous
control~\cite{ref2,ref3,ref5,ref20,ref21}. Recent methods therefore
increasingly pair a VLM for semantic grounding with a dedicated
diffusion or flow-matching action expert for continuous action
generation~\cite{ref6,ref7,ref8,ref9,ref10,ref11,ref22}. While
this design improves action modeling, it also creates a structural
bottleneck: the action expert receives visual information primarily
through compact VLM representations optimized for semantic abstraction,
which can attenuate the precise geometric and temporal
cues that dexterous manipulation demands.

\subsection{Auxiliary Visual Feature Injection}

To mitigate this bottleneck, one line of work enriches the VLM
pathway itself, fusing geometry-grounded or spatially enriched
features into VLM visual representations before semantic
processing, or strengthening the vision backbone
with pretrained geometric and spatiotemporal
priors~\cite{ref23,ref24,ref19,ref25,ref26,ref27,ref28,ref29}. However, because
the enriched signal is still mediated by the VLM, it remains subject
to semantic compression before reaching the action expert. A more
direct approach bypasses the VLM entirely by injecting auxiliary
visual features into the action expert through side branches, inspired
by side-branch conditioning paradigms from conditional
generation~\cite{ref30,ref31,ref32}. Recent VLA variants have
validated this strategy by injecting object-centric or 3D geometric
features into pretrained action experts~\cite{ref14,ref15,ref16,ref17}.
Yet existing methods are largely designed around specific static
encoders and task-specific augmentation, leaving open the
question of how to build a stable, minimally invasive, and
encoder-agnostic injection interface that preserves pretrained
capabilities while enabling systematic comparison across different
visual representations, including temporal ones.

\subsection{Visual Representations for Manipulation: From Static to Temporal}

Existing visual injection methods for VLAs often rely on
static spatial representations, including object-centric features,
point-cloud or geometric cues, and single-frame image embeddings,
which are effective for target localization and scene layout
encoding~\cite{ref14,ref15,ref16,ref17,ref33,ref34,ref35,ref36,ref37,ref39,ref40}.
Yet in fine-grained manipulation, especially bimanual, multi-stage,
and partially observable settings, the policy must also infer motion
progression, alignment convergence, contact transitions, and subtask
completion, signals that are inherently ambiguous from a single
observation. Temporal representations have been studied across video
modeling, robot learning, and VLA models through predictive video
representation learning, recurrent policies, world models, and
video-based conditioning, demonstrating that motion and contact
dynamics provide complementary information beyond static
features~\cite{ref18,ref19,ref25,ref26,ref41,ref42,ref43,ref44,ref45,ref46,ref47}. However, these approaches typically require substantial architectural modifications to accommodate temporal inputs, leaving temporal
representations largely unexplored as lightweight injected conditions
for pretrained action experts. PVI directly targets this gap by
providing an encoder-agnostic injection interface that decouples the
choice of visual representation from the pretrained action expert,
enabling a principled study of temporal dynamics as a complementary
signal for dexterous manipulation.

\section{Method}

\subsection{Preliminaries}

\textbf{VLA Architecture with a Flow-Matching Action Expert.}
We consider a typical VLA architecture composed of two parts: a vision-language model (VLM) backbone for semantic grounding and a Diffusion Transformer (DiT) action expert for continuous action generation.
Given a language instruction $l$ and multi-view image observations $\{I_v\}_{v=1}^{V}$, the VLM jointly encodes them into a sequence of embeddings
\begin{equation}
\mathbf{z}_{\mathrm{vl}} \in \mathbb{R}^{S \times D},
\end{equation}
where $S$ is the sequence length and $D$ is the embedding dimension.

\noindent\textbf{Flow-Matching Action Generation.}
The DiT action expert learns a velocity field $\hat{v}_{\theta}$ that transports Gaussian noise to the target action trajectory.
Let $\mathbf{a}=(\mathbf{a}_1,\ldots,\mathbf{a}_H)$ denote the target action sequence of horizon $H$, and let $\boldsymbol{\epsilon}\sim\mathcal{N}(\mathbf{0},\mathbf{I})$.
At flow time $t\in[0,1]$, the interpolated trajectory is
\begin{equation}
\mathbf{a}_t = (1-t)\boldsymbol{\epsilon} + t\mathbf{a}.
\end{equation}
Conditioned on $\mathbf{z}_{\mathrm{vl}}$, the DiT predicts the velocity field and is trained with the standard flow-matching objective
\begin{equation}
\mathcal{L}=\mathbb{E}_{t,\boldsymbol{\epsilon}}\Big[\big\|\hat{v}_{\theta}(\mathbf{a}_t,t,\mathbf{z}_{\mathrm{vl}})-(\mathbf{a}-\boldsymbol{\epsilon})\big\|_2^2\Big].
\end{equation}
At inference time, the model generates actions by integrating the predicted velocity field from pure noise using $K$-step Euler updates.

\noindent\textbf{The Semantic-Geometric and Temporal Gap.}
In this architecture, $\mathbf{z}_{\mathrm{vl}}$ is the sole channel through which visual information reaches the action expert.
However, VLMs are primarily optimized for high-level semantic understanding, such as language alignment and scene-level reasoning.
As a result, their embeddings do not reliably preserve fine-grained geometric cues (e.g., edges, local depth, contact surfaces) or temporal dynamics (e.g., motion trends and state transitions) that are critical for dexterous manipulation.
This bottleneck is especially severe in bimanual tasks requiring sub-centimeter precision.

To address this, we propose PVI, a lightweight plug-in module 
that injects auxiliary visual representations directly into 
the frozen action expert.

\subsection{PVI: Architecture Overview}

\begin{figure}[tb]
  \centering
  \includegraphics[width=\linewidth]{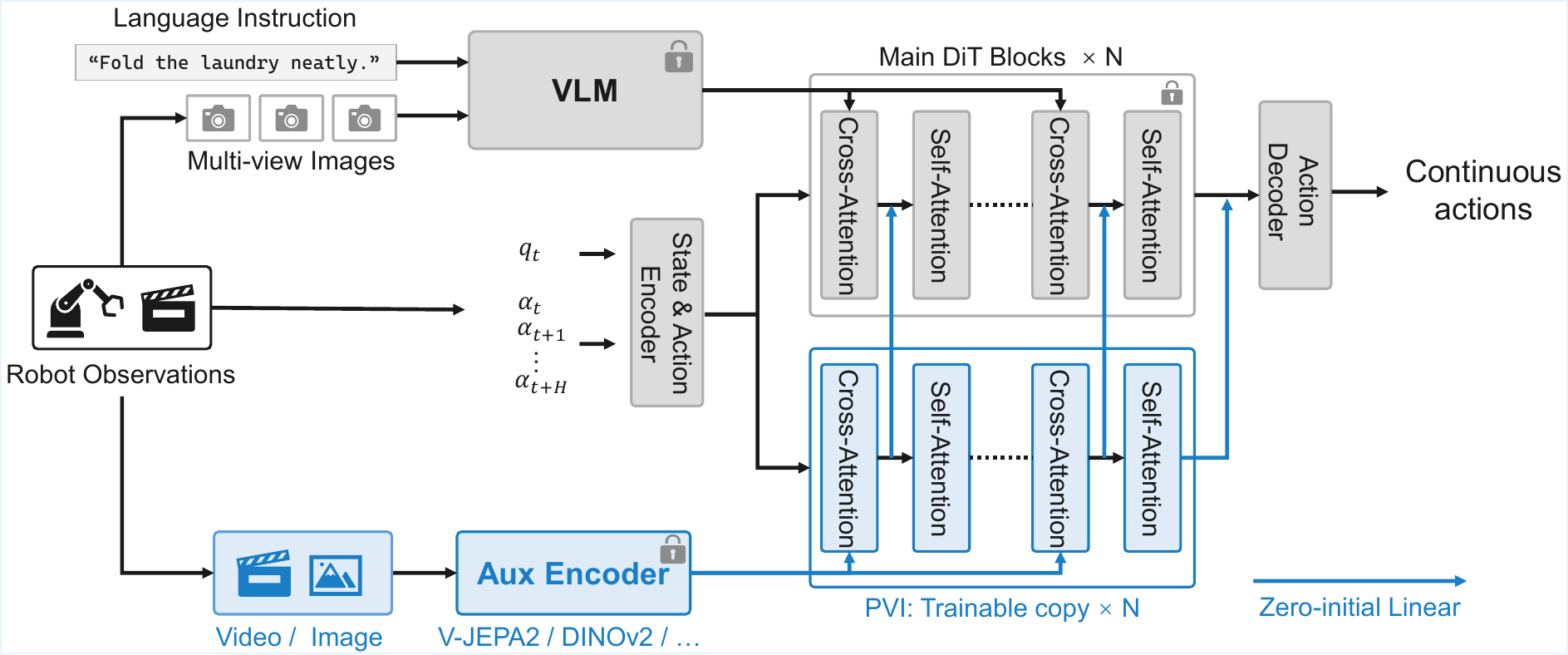}
  \caption{\textbf{Architecture overview.} The frozen main DiT blocks receive semantic embeddings from a frozen VLM. A trainable DiT copy (PVI) conditions on auxiliary visual features and injects them into the main stream via zero-initialized linear projections to produce continuous actions.}
  \label{fig:fig_method}
\end{figure}

PVI augments a pretrained VLA with a side-branch visual injection pathway, while keeping the original architecture and pretrained knowledge intact, as shown in \cref{fig:fig_method}.
The key idea is to inject fine-grained auxiliary visual features directly into the DiT action generation process, rather than relying solely on the VLM embedding.
PVI consists of three components:
\begin{itemize}
  \item \textbf{Encoder-agnostic auxiliary visual feature extraction.}
  A frozen auxiliary visual encoder $E$ extracts features $\mathbf{z}_{\mathrm{aux}}$ from raw observations.
  A trainable zero-initialized projection maps these features into the DiT embedding space.
  PVI imposes no restriction on the choice of $E$, enabling plug-and-play integration of heterogeneous visual representations.

  \item \textbf{Copy-branch DiT with dual conditioning and layer-wise injection.}
  We instantiate a trainable copy branch that mirrors the architecture of
  the main DiT and is initialized from its pretrained weights.
  The copy branch retains the same conditioning mechanism as the main DiT,
  but substitutes the VLM condition $\mathbf{z}_{\mathrm{vl}}$ with the
  projected auxiliary features $\tilde{\mathbf{z}}_{\mathrm{aux}}$.
  Its layer-wise hidden states are injected into the frozen main DiT
  through zero-initialized residual mappings.

  \item \textbf{Zero-initialized training with minimal intervention.}
  We freeze the VLM and the main DiT, and train only the newly introduced projection, copy branch, and injection layers (plus embodiment-specific state/action adapters when applicable).
  Zero initialization ensures that training starts from behavior equivalent to the original pretrained VLA and gradually incorporates auxiliary visual information.
\end{itemize}

We describe these components in detail below.

\subsection{Encoder-Agnostic Auxiliary Visual Feature Extraction}

PVI does not assume a specific auxiliary visual encoder.
We only require an encoder $E$ that takes raw observations as input
and outputs a fixed-dimensional feature sequence
$\mathbf{z}_{\mathrm{aux}} \in \mathbb{R}^{L \times d_E}$,
where $L$ is the auxiliary sequence length and $d_E$ is the encoder
feature dimension.
A trainable linear projection $\mathbf{W}_{\mathrm{proj}} \in \mathbb{R}^{d_E \times D}$
maps these features into the DiT embedding space:
\begin{equation}
\tilde{\mathbf{z}}_{\mathrm{aux}} = \mathbf{z}_{\mathrm{aux}} \mathbf{W}_{\mathrm{proj}}
\in \mathbb{R}^{L \times D}.
\end{equation}
We initialize $\mathbf{W}_{\mathrm{proj}}$ to zero so that the copy branch
receives a zero conditioning signal at the start of training.

This encoder-agnostic design enables controlled comparison across different
classes of visual representations, each compensating for a different
limitation of $\mathbf{z}_{\mathrm{vl}}$.
In this work, we focus on two complementary categories:
\begin{itemize}
  \item \textbf{Temporal dynamic representations,} extracted from video
  clips, which encode motion trajectories and state evolution and
  compensate for the lack of temporal dynamics in static VLM embeddings.

  \item \textbf{Spatial geometric representations,} extracted from image
  observations, which encode boundaries, depth cues, and fine spatial
  structure and compensate for geometric information loss caused by
  semantic compression in the VLM.
\end{itemize}

\subsection{Dual-Pathway DiT with Layer-wise Injection}

\textbf{Shared Initial Features.}
Let the main DiT contain $N$ Transformer blocks.
A proprioceptive state and noisy action tokens are first encoded and
concatenated into an initial token sequence
$\mathbf{h}_0 \in \mathbb{R}^{M \times D}$,
where $M$ is the number of tokens and $D$ is the hidden dimension.
This same $\mathbf{h}_0$ is used as input to both the frozen main DiT
and the trainable copy branch.

\noindent\textbf{Dual Conditioning.}
In a common DiT action expert, each Transformer block receives the VLM
embedding $\mathbf{z}_{\mathrm{vl}}$ through an architecture-dependent conditioning
mechanism (e.g., cross-attention, AdaLN modulation, or conditional
concatenation).
The copy branch preserves the exact architecture and conditioning mechanism
of the main DiT, but replaces $\mathbf{z}_{\mathrm{vl}}$ with
$\tilde{\mathbf{z}}_{\mathrm{aux}}$ as the conditioning input.
This keeps the copy branch structurally aligned with the pretrained action
expert while specializing it to process auxiliary visual information.

For example, GR00T N1.5 uses cross-attention layers conditioned on
$\mathbf{z}_{\mathrm{vl}}$ alongside self-attention.
In the PVI copy branch, the same cross-attention structure is retained,
but its key/value inputs are replaced by $\tilde{\mathbf{z}}_{\mathrm{aux}}$.
The same principle applies to other conditioning mechanisms.

\noindent\textbf{Layer-Wise Residual Injection.}
Denote the $i$-th block of the frozen main DiT by $f_i^{\mathrm{main}}$
and the corresponding block in the trainable copy branch by
$f_i^{\mathrm{copy}}$.
The copy branch is computed as
\begin{equation}
\mathbf{h}_i^{\mathrm{copy}} = f_i^{\mathrm{copy}}\!\left(
\mathbf{h}_{i-1}^{\mathrm{copy}}, \tilde{\mathbf{z}}_{\mathrm{aux}}\right),
\quad i=1,\ldots,N.
\end{equation}
The main DiT is computed in parallel and receives a layer-wise control
signal from the copy branch:
\begin{equation}
\mathbf{h}_i^{\mathrm{main}} =
f_i^{\mathrm{main}}\!\left(\mathbf{h}_{i-1}^{\mathrm{main}},
\mathbf{z}_{\mathrm{vl}}\right)
+ \mathbf{Z}_i\!\left(\mathbf{h}_i^{\mathrm{copy}}\right),
\quad i=1,\ldots,N,
\end{equation}
where $\mathbf{Z}_i$ is a zero-initialized linear injection layer.
This design allows auxiliary visual information to influence action
generation at every layer of the DiT while preserving the pretrained
main pathway.

\subsection{Training Strategy}
We freeze all pretrained components of the original VLA, including the
VLM backbone and the main DiT, as well as the auxiliary visual encoder $E$.
The trainable parameters are the projection layer $\mathbf{W}_{\mathrm{proj}}$,
the copy-branch blocks $\{f_i^{\mathrm{copy}}\}_{i=1}^{N}$,
the injection layers $\{\mathbf{Z}_i\}_{i=1}^{N}$,
and embodiment-specific state/action encoders and decoders that bridge the target
embodiment to the pretrained VLA interface.

The copy branch is initialized by copying the pretrained weights of the
main DiT, providing a strong feature-processing prior from the start of
training.
The projection layer $\mathbf{W}_{\mathrm{proj}}$ and injection layers
$\{\mathbf{Z}_i\}_{i=1}^{N}$ are initialized to zero, which guarantees
functional equivalence to the original pretrained VLA at initialization:
the auxiliary pathway produces zero control signals and does not perturb
the frozen main pathway.
As training proceeds, the injection layers learn non-zero mappings,
enabling gradual integration of auxiliary visual information and reducing
the risk of interference with pretrained knowledge.

PVI uses the same flow-matching loss as the baseline and introduces no
auxiliary supervision.
All newly introduced parameters are trained solely through the action
prediction objective.

\section{Experiments}

To validate PVI, we instantiate it on GR00T N1.5 using the released
pretrained model as our base.
We begin with a quantitative comparison of candidate injection strategies
in simulation to evaluate whether PVI provides a stronger mechanism for
delivering auxiliary visual information to the flow-matching action expert
(\cref{sect:comparing_injection}).
We then demonstrate encoder-agnostic generality by plugging in different
auxiliary encoders and conducting a controlled representation study that
compares temporal and spatial visual features, as well as their combinations
(\cref{sect:Encoder-Agnostic}).
Next, we analyze key design choices through ablations on temporal context
length and stabilization options (\cref{sect:Ablations}), and evaluate PVI
under a more challenging multi-task regime (\cref{sect:Multi-Task}). Finally, we provide a qualitative real-robot demonstration on bimanual
cloth folding, a long-horizon task requiring precise two-arm coordination
over a deformable object, to validate the practicality of deploying PVI beyond
simulation (\cref{sect:Real-World}).

Across all simulation experiments, we enforce a consistent training and
evaluation protocol for fair comparison.
All methods, including the baseline GR00T N1.5, GR00T N1.5 + PVI, and
other candidate injection strategies, start from the same pretrained
weights and undergo a single round of fine-tuning on the same task data,
with identical training steps, learning rate, batch size, and all other
hyperparameters.
The only difference across conditions is the injection architecture and
the associated freezing strategy. Full implementation details for all variants (architectures, freezing policies, trainable parameter counts) and training hyperparameters are provided in the supplementary material.

\subsection{Comparing Injection Strategies}
\label{sect:comparing_injection}

\begin{figure}[tb]
  \centering
  \includegraphics[width=\linewidth]{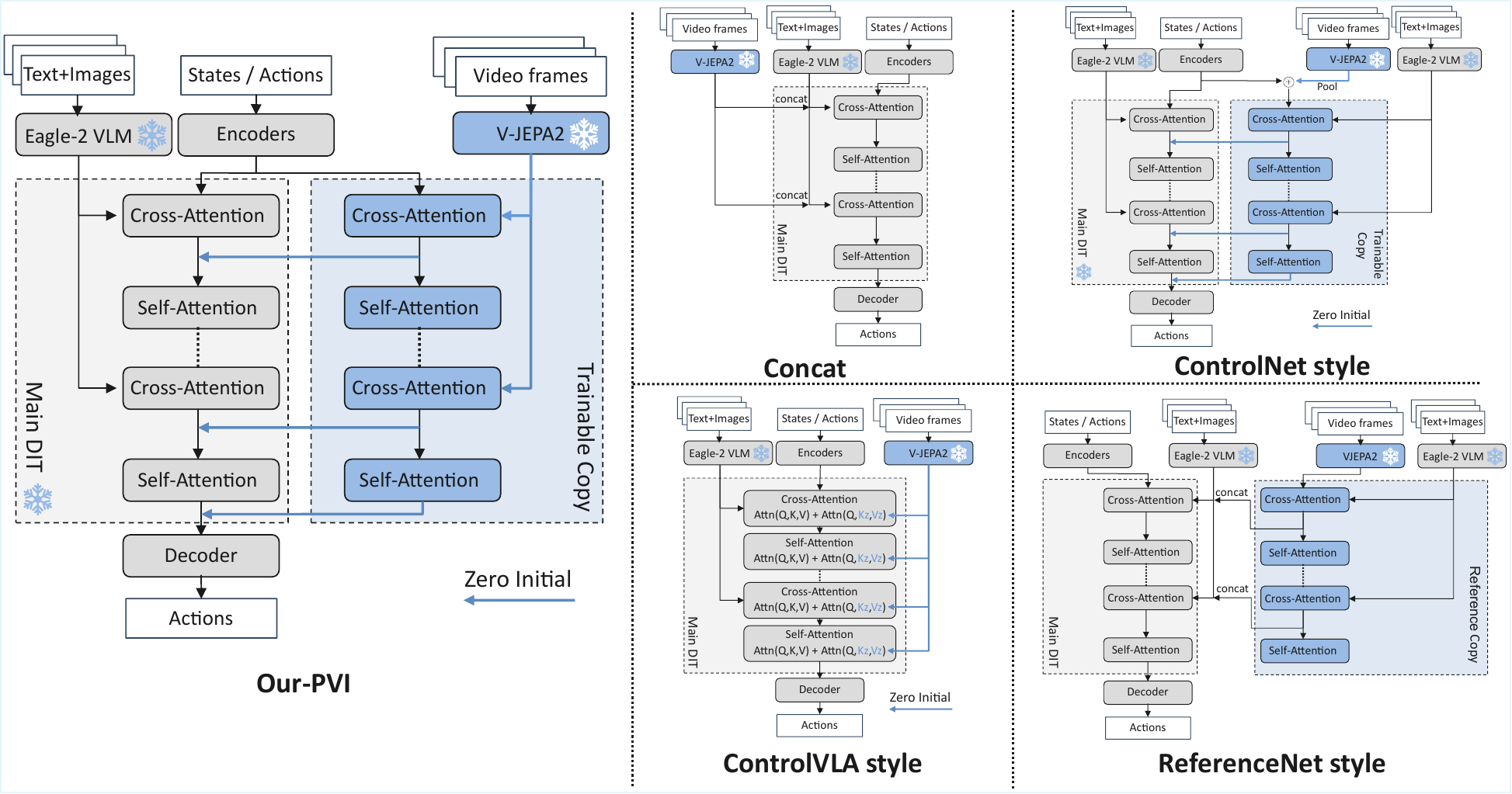}
\caption{\textbf{Overview of PVI and four candidate strategies for 
injecting auxiliary visual features into the DiT action expert.} 
We compare \textbf{PVI} (ours), which injects V-JEPA2 features via 
a trainable copy branch with zero-initialized injection layers, 
against input-level fusion (\textbf{Concat}), attention-level 
dual injection (\textbf{ControlVLA-style}), and parallel-branch 
designs with feature concatenation or residual addition 
(\textbf{ReferenceNet-} and \textbf{ControlNet-style}).}
  \label{fig:injection_methods}
\end{figure}

\begin{table*}[tb]
  \caption{\textbf{Comparing injection strategies in simulation
(single-task fine-tuning).}
Success rates (\%) over 100 evaluation rollouts per task with randomized initial conditions (different rollout seeds).
Evaluated tasks span diverse manipulation primitives, including
alignment and perception, press and switch, relocation and placement,
articulation, and coordination with tools and stacking.}
  \label{tab:inject_strategies}
  \centering
  \scriptsize
  \setlength{\tabcolsep}{4pt}
  \renewcommand{\arraystretch}{1.05}
  \begin{tabular*}{\textwidth}{@{\extracolsep{\fill}}lcccccc@{}}
    \toprule
    Task & GR00T N1.5 & Ours & Concat & ControlNet & ControlVLA & ReferenceNet \\
    \midrule
    \texttt{adjust\_bottle}        & 92 & 95 & 90 & 86 & 91 & \cellcolor{black!6}\textbf{96} \\
    \texttt{beat\_block\_hammer}    & 39 & \cellcolor{black!6}\textbf{84} & 47 & 30 & 35 & 37 \\
    \texttt{click\_alarmclock}      & 48 & \cellcolor{black!6}\textbf{87} & 54 & 46 & 53 & 48 \\
    \texttt{move\_can\_pot}         & 37 & \cellcolor{black!6}\textbf{51} & 36 & 30 & 37 & 37 \\
    \texttt{click\_bell}            & 46 & \cellcolor{black!6}\textbf{91} & 75 & 41 & 61 & 42 \\
    \texttt{move\_playingcard\_away} & 23 & \cellcolor{black!6}\textbf{84} & 39 & 29 & 26 & 61 \\
    \texttt{pick\_diverse\_bottles} &  3 & \cellcolor{black!6}\textbf{24} &  4 & 10 &  4 &  5 \\
    \texttt{place\_a2b\_left}       & 31 & \cellcolor{black!6}\textbf{36} & 31 & 21 & 28 & 19 \\
    \texttt{open\_laptop}           & 48 & 70 & \cellcolor{black!6}\textbf{71} & 38 & 53 & 61 \\
    \texttt{press\_stapler}         & 31 & \cellcolor{black!6}\textbf{72} & 43 & 52 & 31 & 41 \\
    \texttt{rotate\_qrcode}         & 38 & \cellcolor{black!6}\textbf{55} & 45 & 26 & 36 & 46 \\
    \texttt{scan\_object}           &  4 & \cellcolor{black!6}\textbf{42} & 14 & 18 & 10 & 15 \\
    \texttt{put\_object\_cabinet}   & 24 & \cellcolor{black!6}\textbf{36} & 18 & 22 & 23 & 21 \\
    \texttt{turn\_switch}           & 28 & \cellcolor{black!6}\textbf{62} & 25 & 57 & 27 & 27 \\
    \texttt{handover\_block}        & 21 & \cellcolor{black!6}\textbf{50} & 45 & 27 & 24 & 28 \\
    \texttt{open\_microwave}        & 38 & 51 & \cellcolor{black!6}\textbf{54} & 34 & 46 & 49 \\
    \texttt{stack\_blocks\_two}     & 40 & \cellcolor{black!6}\textbf{60} & 49 & 19 & 25 & 23 \\
    \texttt{stack\_bowls\_two}      & 79 & 86 & 85 & 76 & \cellcolor{black!6}\textbf{89} & 73 \\
    \texttt{blocks\_ranking\_rgb}   & 33 & \cellcolor{black!6}\textbf{43} & 32 & 21 & 34 & 12 \\
    \texttt{blocks\_ranking\_size}  & 11 & \cellcolor{black!6}\textbf{15} & \cellcolor{black!6}\textbf{15} &  4 &  7 &  7 \\
    \midrule
    \textbf{Average success} & 35.70 & \cellcolor{black!6}\textbf{59.70} & 43.60 & 34.35 & 37.00 & 37.40 \\
    $\Delta$ vs. Base (pp) & -- & \cellcolor{black!6}\textbf{$+$24.00} & $+$7.90 & $-$1.35 & $+$1.30 & $+$1.70 \\
    \bottomrule
  \end{tabular*}
\end{table*}

We compare PVI against several candidate strategies for injecting 
auxiliary visual information into a pretrained flow-matching action 
expert. We draw inspiration from related domains to instantiate 
four plausible injection designs: Concat represents a natural 
baseline; ControlNet- and ReferenceNet-style strategies adapt 
conditioning mechanisms from text-to-image generation~\cite{ref32,ref31}; 
and ControlVLA-style adapts the zero-initialized dual cross-attention 
mechanism from ControlVLA~\cite{ref14}, which was originally 
proposed for injecting object-centric features into pretrained VLA 
models. To isolate the effect of the injection mechanism, we fix 
the auxiliary encoder to V-JEPA2 and the training budget 
across all methods and vary only how auxiliary features are 
introduced into the action expert (\cref{fig:injection_methods}).

We consider four alternative designs:

(1) \textbf{Concat-style} concatenates auxiliary visual embeddings 
with VLM representations as the joint cross-attention key/value context for 
the DiT. Since the action expert must learn to interpret the 
expanded context, the full DiT is fine-tuned end-to-end.

(2) \textbf{ControlNet-style} maintains a trainable copy 
of the frozen action expert, where auxiliary visual features are 
added to the state-action embeddings before entering the control 
branch. The control branch also receives VLM representations as 
cross-attention conditioning, and injects its intermediate 
representations into the frozen main DiT via zero-initialized 
residual additions at each block.

(3) \textbf{ReferenceNet-style} similarly uses a parallel 
DiT copy, but replaces the state-action embedding input entirely 
with auxiliary visual features while retaining VLM conditioning. 
Unlike ControlNet-style, both the main DiT and the reference branch 
are trained jointly, incurring higher training cost.

(4) \textbf{ControlVLA-style} injects auxiliary 
features directly at the attention computation level: an auxiliary 
attention term $\text{softmax}(QK_z^T/\sqrt{d})V_z$ is added 
alongside the standard attention, where $K_z, V_z$ are 
zero-initialized projections of the auxiliary features. The full 
DiT and auxiliary projections are optimized jointly.


We conduct a quantitative evaluation in RoboTwin 2.0\cite{chen2025robotwin}, a bimanual manipulation
simulator covering diverse task types and manipulation primitives.
Following the shared protocol described above, we fine-tune each method
independently on each task using 50 demonstration trajectories and report
success rates over 100 evaluation rollouts with randomized initial conditions.
Unless otherwise stated, all methods use a frozen V-JEPA2 encoder taking
8 frames sampled at 4\,fps as the auxiliary feature source.

\cref{tab:inject_strategies} summarizes results across 20 bimanual tasks.
The fine-tuned GR00T N1.5 baseline achieves \textbf{35.7\%} average success, while
GR00T N1.5 + PVI reaches \textbf{59.7\%} (\textbf{+24.0} pp).
Concat provides a moderate improvement (\textbf{43.6\%}), whereas the ControlVLA-,
ControlNet-, and ReferenceNet-style variants remain close to the baseline
(\textbf{34.4--37.4\%}).
PVI's gains are particularly pronounced on tasks with low baseline success,
for example \texttt{beat\_block\_hammer} (\textbf{39\%} $\to$ \textbf{84\%}),
\texttt{click\_alarmclock} (\textbf{48\%} $\to$ \textbf{87\%}),
\texttt{move\_playingcard\_away} (\textbf{23\%} $\to$ \textbf{84\%}),
and \texttt{scan\_object} (\textbf{4\%} $\to$ \textbf{42\%}).
These results support two conclusions: injecting temporal evidence into the
action expert is broadly beneficial for contact-rich bimanual manipulation,
and the injection mechanism itself matters significantly.
PVI's dual-conditioning and behavior-preserving layer-wise injection yield
substantially larger gains than competing designs under the same data and
training budget.

\subsection{Encoder-Agnostic Injection and Representation Study}
\label{sect:Encoder-Agnostic}

\begin{figure}[tb]
  \centering
  \includegraphics[width=\linewidth]{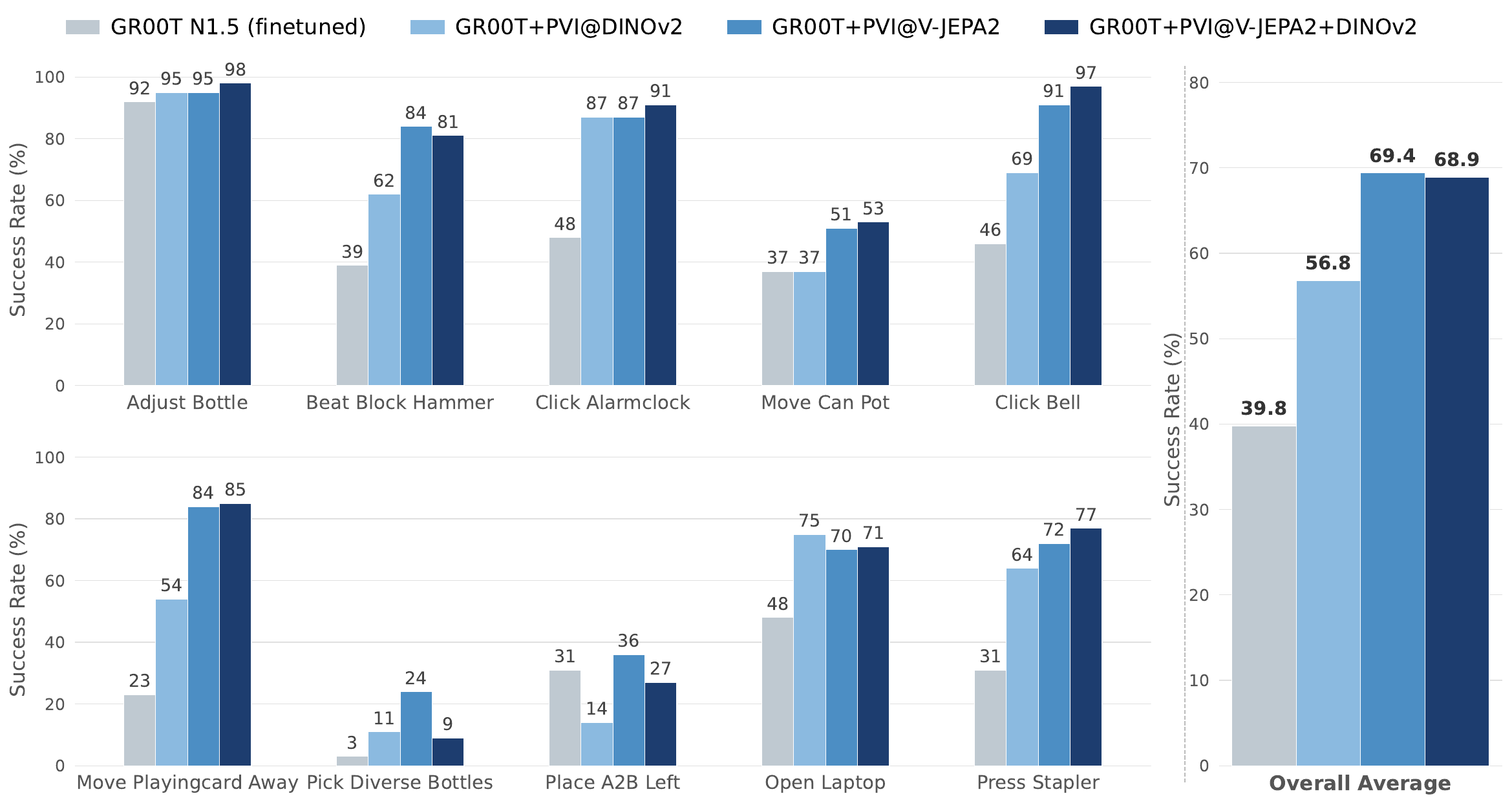}
  \caption{\textbf{Encoder-agnostic visual injection and representation
comparison.}
Per-task success rates and overall average across 10 simulation tasks,
comparing a fine-tuned GR00T N1.5 baseline against three PVI
instantiations with different auxiliary encoders.}
  \label{fig:Encoder-Agnostic}
\end{figure}

Having established that the injection mechanism matters, we next study what auxiliary visual information should be injected. A key advantage of PVI is its encoder-agnostic interface: the same injection pathway can accept heterogeneous auxiliary encoders without architectural changes to the pretrained action expert. This enables two goals under a unified setup: (i) validating that PVI can plug in different representations and consistently improve a strong base VLA, and (ii) conducting a controlled comparison where the injected representation is the primary variable.

We evaluate on 10 simulation tasks from RoboTwin 2.0 following the same protocol described in Sec.~\ref{sect:comparing_injection}.
We denote GR00T augmented with PVI and an auxiliary encoder $E$ as
GR00T+PVI@$E$, and compare three instantiations:
GR00T+PVI@DINOv2, which injects static image features from a single frame;
GR00T+PVI@V-JEPA2, which injects temporal video features from 8 frames
sampled at 4\,fps; and GR00T+PVI@V-JEPA2+DINOv2, which combines both
encoders through the same interface.

Results are shown in \cref{fig:Encoder-Agnostic}.
The fine-tuned GR00T N1.5 baseline achieves \textbf{39.8\%} average success.
All three PVI instantiations consistently improve over the baseline,
confirming that the encoder-agnostic interface transfers benefit regardless
of the encoder type.
Injecting static features already yields a substantial gain:
GR00T+PVI@DINOv2 reaches \textbf{56.8\%}
(\textbf{+17.0} pp, \textbf{+42.7\%} relative).
Temporal features provide the largest improvement:
GR00T+PVI@V-JEPA2 reaches \textbf{69.4\%}
(\textbf{+29.6} pp, \textbf{+74.4\%} relative).
Combining both encoders achieves \textbf{68.9\%}
(\textbf{+29.1} pp, \textbf{+73.1\%} relative), on par with
V-JEPA2 alone, suggesting that static appearance features provide
little additional signal beyond what temporal features already capture.
The consistent advantage of temporal features across tasks, particularly
those involving multi-phase coordination and state tracking, points to
temporally coherent evidence as the more critical ingredient for
closed-loop bimanual manipulation.

\subsection{Ablations and Sensitivity Analysis}
\label{sect:Ablations}
We analyze the sensitivity of PVI to temporal context length and
stabilization/adaptation choices. All ablations follow the same protocol and use 10 simulation tasks from
RoboTwin 2.0 as in \cref{sect:Encoder-Agnostic}, reporting average
success over 100 rollouts per task. Results are summarized in \cref{tab:ablation}.

Varying the number of frames fed to the frozen V-JEPA2 encoder at 4\,fps,
we find that a moderate temporal context is sufficient: 2--4 frames achieve
the best performance (\textbf{71.6\%}--\textbf{71.8\%}), while longer contexts
yield diminishing returns (\textbf{69.4\%} at 8 frames, \textbf{66.6\%} at
16 frames). Notably, the default 8-frame setting used throughout our main
experiments is not the optimal configuration identified here; the performance
gap is small (2.4 pp) but suggests that the gains reported in
Sec.~\ref{sect:comparing_injection} and Sec.~\ref{sect:Encoder-Agnostic}
are conservative, and that PVI can be further improved with task-specific
tuning of temporal context. Combining temporal and static encoders through
the same interface yields competitive performance (\textbf{68.9\%} with
V-JEPA2+DINOv2 at 16 frames), suggesting limited additional benefit beyond
temporal features alone.

We further ablate stabilization and adaptation choices. Freezing the
auxiliary-to-DiT projector substantially reduces performance (\textbf{55.3\%}),
indicating that learning feature alignment into the action expert's embedding
space is important even when the auxiliary encoder is frozen. Removing zero
initialization yields comparable or slightly higher final success
(\textbf{71.8\%}); nevertheless, we retain zero initialization in our main
experiments because it guarantees behavior-preserving initialization and
supports conservative, progressive integration of auxiliary signals during
fine-tuning.

\begin{table}[tb]
  \caption{\textbf{Ablations on temporal context and stabilization choices.}
  Average success rates (\%) over 100 evaluation rollouts per task with randomized initial conditions (different rollout seeds).}
  \label{tab:ablation}
  \centering
  \small
  \renewcommand{\arraystretch}{1.05}
  \begin{tabular*}{0.9\linewidth}{@{\extracolsep{\fill}}lcr@{}}
    \toprule
    Variant & Frames@4fps & Avg. (\%) \\
    \midrule
    Baseline (fine-tuned GR00T N1.5) & -- & 39.8 \\
    \midrule
    \multicolumn{3}{@{}l@{}}{\textit{Temporal context (PVI@V-JEPA2)}}\\
    PVI@V-JEPA2 & 2  & 71.6 \\
    PVI@V-JEPA2 & 4  & 71.8 \\
    PVI@V-JEPA2 & 8  & 69.4 \\
    PVI@V-JEPA2 & 16 & 66.6 \\
    PVI@V-JEPA2 + DINOv2 & 16 & 68.9 \\
    \midrule
    \multicolumn{3}{@{}l@{}}{\textit{Stabilization / adaptation (PVI@V-JEPA2, 8f)}}\\
    PVI@V-JEPA2 (no zero-init) & 8 & 71.8 \\
    PVI@V-JEPA2 (freeze projector) & 8 & 55.3 \\
    \bottomrule
  \end{tabular*}
\end{table}

\subsection{Multi-Task Scalability}
\label{sect:Multi-Task}
We further evaluate PVI in a multi-task regime, where a single 
policy must cover a broad set of skills while sharing the same 
action expert across tasks. We train on a 20-task and a 50-task 
RoboTwin 2.0 mixture with 50 demonstrations per task, starting 
from the same pretrained GR00T N1.5 checkpoint. We report 
per-task success rates over 100 rollouts and summarize 
performance by average success rate.

PVI consistently improves over the fine-tuned GR00T N1.5 baseline 
in both settings. On the 20-task mixture, PVI increases average 
success from 61.15\% to 69.15\% (+8.00 pp). On the more 
challenging 50-task mixture, PVI yields a consistent gain from 
61.32\% to 63.56\% (+2.24 pp), winning on the majority of 
individual tasks. These results suggest that auxiliary visual 
injection remains beneficial even when training a shared policy 
over diverse manipulation primitives. Full per-task results are 
provided in the supplementary material.

\subsection{Real-World Bimanual Deformable-Object Demonstration}
\label{sect:Real-World}
\begin{figure}[tb]
  \centering
  \includegraphics[width=\linewidth]{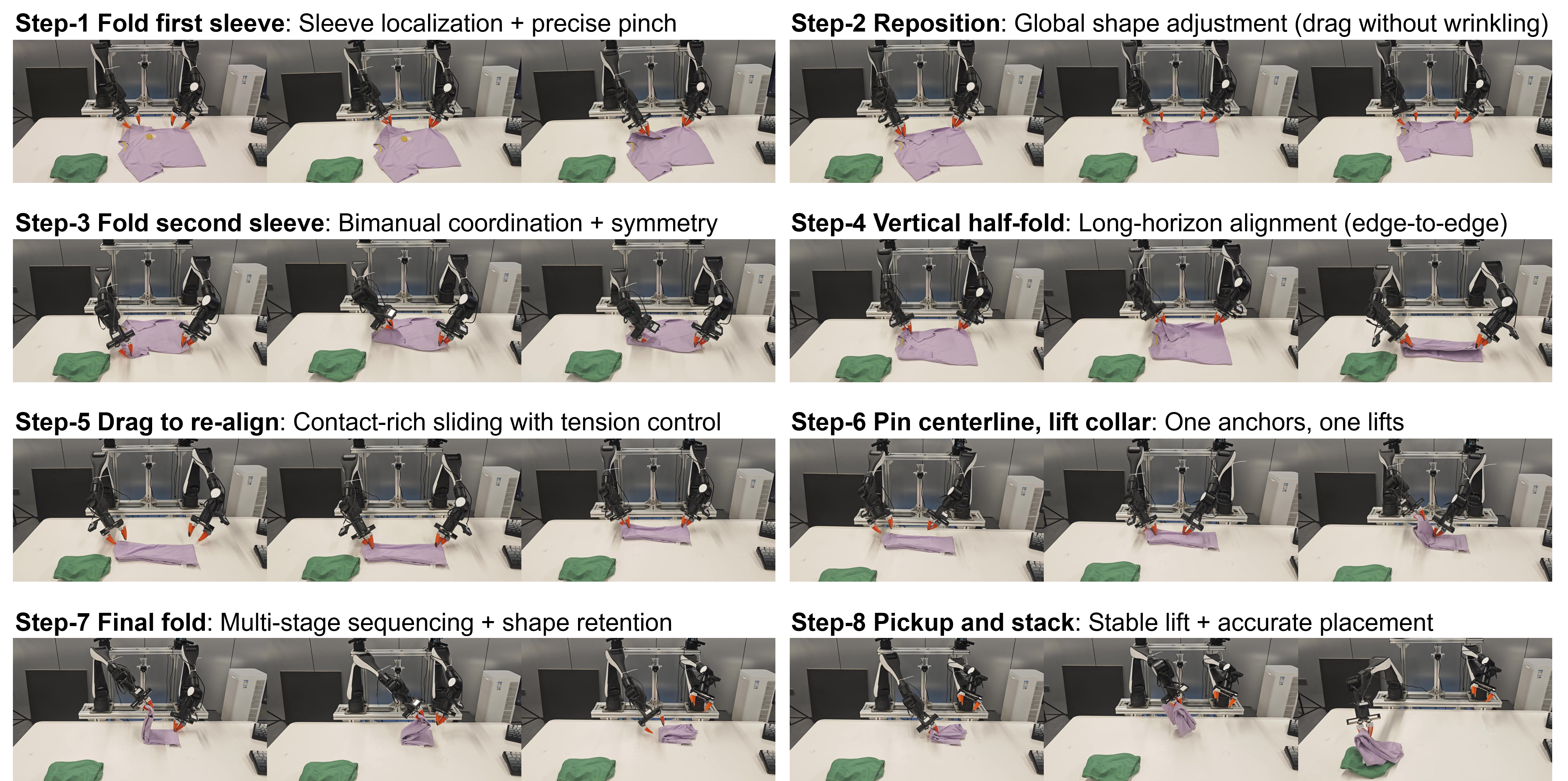}
  \caption{\textbf{PVI enables long-horizon bimanual manipulation of deformable objects on real hardware.} The task
  comprises eight sequential subtasks, each illustrated by three keyframes:
  sleeve localization and precise pinching (Step~1), global shape adjustment
  via dragging without wrinkling (Step~2), symmetric bimanual sleeve folding
  (Step~3), long-horizon edge-to-edge vertical half-fold (Step~4),
  contact-rich sliding with tension control (Step~5), asymmetric anchoring
  and collar lifting (Step~6), multi-stage final fold with shape retention
  (Step~7), and stable lift-and-stack (Step~8). All subtasks are driven by
  a single PVI-augmented policy conditioned on language instructions, without
  task-specific engineering or manual resets between steps.}
  \label{fig:real_demo}
\end{figure}
To assess the practicality of PVI beyond simulation, we deploy it on an
Airbot dual-arm platform for long-horizon bimanual cloth folding---a task
that demands precise two-arm coordination, contact-rich interaction, and
sustained shape reasoning over a highly deformable garment. We collect a set of demonstration trajectories and fine-tune the GR00T N1.5
checkpoint with PVI attached, following the same single-stage fine-tuning
procedure as in simulation without any task-specific engineering.

The resulting policy executes a full eight-stage folding sequence in closed
loop, driven by language instructions without manual resets between subtasks.
As shown in \cref{fig:real_demo}, the policy handles the diverse manipulation
primitives required by the task---precise pinching, global shape adjustment,
symmetric bimanual coordination, and contact-rich sliding with tension
control---across all eight stages. This demonstration provides qualitative evidence that PVI can be fine-tuned and deployed on real bimanual hardware and remains effective for challenging long-horizon deformable-object manipulation.

\section{Conclusion}
We presented Plug-in Visual Injection (PVI), a lightweight,
encoder-agnostic framework that injects auxiliary visual representations
directly into the frozen action expert of a pretrained VLA via a
behavior-preserving, layer-wise residual pathway, enabling stable
single-stage fine-tuning without any backbone modification. Across a
broad suite of bimanual manipulation tasks, PVI substantially improves
over a fine-tuned GR00T N1.5 baseline and outperforms a range of
competitive injection strategies, demonstrating that \emph{how}
auxiliary information is injected is as important as \emph{what} is
injected. Using PVI as a controlled testbed, we further show that
temporal video representations outperform strong static image features
for fine-grained multi-phase manipulation, with the largest gains on
tasks requiring state tracking and cross-arm coordination. PVI scales to
multi-task regimes and transfers to real bimanual hardware for
long-horizon deformable-object manipulation, establishing a simple and
stable foundation for integrating richer visual signals into future
VLA systems.

\section*{Acknowledgements}
We thank Dong Sun and Chaorong Zhang for their contributions to 
bimanual real-robot data collection and processing, and Ting Sun 
for his contributions to the training framework.

%
%
\bibliographystyle{splncs04}
\bibliography{main}

\clearpage
\appendix
\section{Implementation Details of All Variants}
\label{sec:supp_impl}

This section summarizes the implementation details of the baseline and all compared injection variants, organized into architectures, training setup, and cost-related statistics.

\subsection{architectures}

\paragraph{Baseline and shared policy interface.}
All variants share the same GR00T-N1.5 policy interface. The policy takes RGB observations from three views (an environment camera, a left wrist camera, and a right wrist camera) together with the robot state, and predicts an action chunk of horizon 16. In the standard GR00T pathway, the visual input consists of the current frame only from each view, with raw resolution $480\times640\times3$ per image, resized to the input resolution required by the pretrained language-vision backbone. The state input also uses the current timestep only. Both state and action are represented in joint space rather than end-effector pose: each is 14-dimensional, comprising 6 arm-joint variables and 1 gripper variable for each arm. In the policy interface, state and action are padded to 64 and 32 dimensions, respectively. The baseline uses this standard pipeline without any auxiliary V-JEPA2 pathway.

\paragraph{V-JEPA2 encoder.}
For the injection-method comparison, the auxiliary visual encoder is a frozen V-JEPA2 model. Its input consists of temporally sampled history frames from three camera views; when the sampled history extends beyond the valid episode range, out-of-range indices are padded by repeating the nearest available frame. For each sample, the raw video is organized as $[V,T,480,640,3]$, where $V=3$ is the number of views and $T$ is the number of sampled history frames. The video is converted to floating point in $[0,1]$, resized following the V-JEPA2 preprocessing rule (short side resized to 438), center-cropped to $384\times384$, and normalized using ImageNet statistics. After preprocessing, the encoder input is arranged as $[B,V,C,T,384,384]$ and reshaped to $[B\!\times\!V,C,T,384,384]$ before being fed into the frozen V-JEPA2 encoder. The encoder outputs patch-level visual features of shape $[B\!\times\!V,N_{\text{patches}},1408]$. In the main comparison of Sec.~4.1, we set $T=8$ and sample frames at 4 FPS.

\paragraph{DINOv2 encoder.}
For the encoder comparison in Sec.~4.2, we replace V-JEPA2 with a frozen DINOv2-Giant image encoder. Given the preprocessed visual tensor $[B,V,C,T,384,384]$, we extract the current frame (i.e., the last frame along the temporal dimension) to form $[B,V,C,384,384]$, reshape it to $[B\!\times\!V,C,384,384]$, and feed it into the frozen DINOv2-Giant encoder. We use the patch tokens from the last hidden state, excluding the CLS token, which yields features of shape $[B\!\times\!V,N_{\text{patches}},1536]$. These features are reshaped to $[B,V\!\times\!N_{\text{patches}},1536]$ for subsequent projection and injection.

\paragraph{Implementation differences across injection variants.}
Across the compared injection variants, all methods use an additional projector to map auxiliary features into the action-expert interface, but the target dimension depends on the injection design. PVI, Concat, and ControlVLA-style project the auxiliary features from 1408 to the 2048-dimensional cross-attention space, whereas ControlNet-style and ReferenceNet-style project them to the 1536-dimensional DiT hidden space. PVI, ControlNet-style, and ReferenceNet-style additionally introduce a copied DiT-based branch initialized from the pretrained main DiT, while Concat and ControlVLA-style do not use an extra copied branch. The main DiT remains frozen in PVI and ControlNet-style, but is jointly trained in Concat, ReferenceNet-style, and ControlVLA-style.

\subsection{Training Setup}

Unless otherwise specified, all experiments are initialized from the same pretrained GR00T-N1.5-3B checkpoint. For the comparisons in Secs.~4.1--4.3, we follow the GR00T recommended fine-tuning schedules, using learning rate $1\times10^{-4}$ and batch size 32 on 4 NVIDIA H20Z GPUs. Standard tasks are trained for 20k steps, while longer-horizon tasks are trained for 60k steps. Within each comparison, the baseline and all compared injection variants use exactly the same training schedule.

For the more challenging multi-task and real-robot experiments in Secs.~4.4--4.5, we use a longer training schedule of about 60 epochs with learning rate $1\times10^{-4}$ and batch size 64 on 8 NVIDIA H20Z GPUs.

\subsection{Cost-Related Statistics}

Tables~\ref{tab:param_main_part1}, \ref{tab:param_main_part2}, and \ref{tab:param_encoder} report module-wise parameter breakdowns for the main injection-method comparison and the encoder comparison. We distinguish \emph{trainable} and \emph{total} parameter counts, since they reflect different costs: the former is more directly related to the optimization budget, while the latter also includes frozen auxiliary modules.

For the main injection-method comparison, PVI keeps the trainable budget comparable to, and in our case lower than, standard GR00T-N1.5 fine-tuning (901.73M vs.\ 1068.81M), even though it introduces an auxiliary encoder and an additional plug-in branch. This is because PVI freezes the original main DiT and places the newly introduced trainable capacity in the plug-in pathway, rather than increasing the optimized portion of the pretrained backbone. Importantly, PVI is trained in a single stage directly from the pretrained GR00T-N1.5 checkpoint, without requiring a separate ``first fine-tune GR00T-N1.5, then fine-tune PVI'' procedure. Therefore, in terms of trainable optimization budget, PVI does not require a larger fine-tuning budget than the standard baseline, while providing substantial performance gains.

By contrast, variants that continue to fine-tune the original main DiT, such as Concat and ControlVLA-style, remain at baseline-level trainable cost, while ReferenceNet-style becomes substantially heavier because it jointly trains both the main DiT and an additional copied branch. In this sense, the parameter increase introduced by PVI should be interpreted mainly as the cost of attaching an external visual pathway, rather than of enlarging the trainable core of the original policy.

A similar pattern appears in the encoder comparison. Under the same PVI framework, replacing V-JEPA2 with DINOv2 or with the dual-encoder combination changes the trainable budget only marginally (901.73M, 901.99M, and 904.88M, respectively). The main increase is in total parameters, which is dominated by the frozen auxiliary encoder(s), while the trainable plug-in pathway remains nearly unchanged. This indicates that the cost difference across encoder choices primarily comes from the frozen visual frontend, rather than from a substantially larger optimized policy.

\begin{table}[htbp]
\centering
\small
\caption{\textbf{Parameter breakdown for the main injection-method comparison in Sec.~4.1 (Part I).}
All numbers are in billions of parameters (B).}
\label{tab:param_main_part1}
\setlength{\tabcolsep}{3.2pt}
\begin{tabular}{
p{2.3cm}
S[table-format=1.3]@{\,$|$\,}S[table-format=1.3]
S[table-format=1.3]@{\,$|$\,}S[table-format=1.3]
S[table-format=1.3]@{\,$|$\,}S[table-format=1.3]
}
\toprule
& \multicolumn{2}{c}{\textbf{Baseline}}
& \multicolumn{2}{c}{\textbf{PVI}}
& \multicolumn{2}{c}{\textbf{Concat}} \\
\cmidrule(lr){2-3} \cmidrule(lr){4-5} \cmidrule(lr){6-7}
\textbf{Quantity}
& \multicolumn{1}{c}{\textbf{Trainable}} & \multicolumn{1}{c}{\textbf{Total}}
& \multicolumn{1}{c}{\textbf{Trainable}} & \multicolumn{1}{c}{\textbf{Total}}
& \multicolumn{1}{c}{\textbf{Trainable}} & \multicolumn{1}{c}{\textbf{Total}} \\
\midrule
VLM
& 0.000 & 1.655
& 0.000 & 1.655
& 0.000 & 1.655 \\

Main DiT
& 0.550 & 0.550
& 0.000 & 0.550
& 0.550 & 0.550 \\

Others
& 0.518 & 0.518
& 0.317 & 0.518
& 0.518 & 0.518 \\

\midrule
Aux encoder
& 0.000 & 0.000
& 0.000 & 1.012
& 0.000 & 1.012 \\

Plug-in branch
& 0.000 & 0.000
& 0.585 & 0.591
& 0.003 & 0.003 \\

\midrule
Total model
& 1.069 & 2.724
& \textbf{0.902} & 4.327
& 1.072 & 3.739 \\

Trainable ratio
& \multicolumn{2}{c}{39.23\%}
& \multicolumn{2}{c}{\textbf{20.84\%}}
& \multicolumn{2}{c}{28.66\%} \\
\bottomrule
\end{tabular}
\end{table}

\begin{table}[htbp]
\centering
\small
\caption{\textbf{Parameter breakdown for the main injection-method comparison in Sec.~4.1 (Part II).}
All numbers are in billions of parameters (B).}
\label{tab:param_main_part2}
\setlength{\tabcolsep}{3.2pt}
\begin{tabular}{
p{2.3cm}
S[table-format=1.3]@{\,$|$\,}S[table-format=1.3]
S[table-format=1.3]@{\,$|$\,}S[table-format=1.3]
S[table-format=1.3]@{\,$|$\,}S[table-format=1.3]
}
\toprule
& \multicolumn{2}{c}{\textbf{ControlVLA-style}}
& \multicolumn{2}{c}{\textbf{ControlNet-style}}
& \multicolumn{2}{c}{\textbf{ReferenceNet-style}} \\
\cmidrule(lr){2-3} \cmidrule(lr){4-5} \cmidrule(lr){6-7}
\textbf{Quantity}
& \multicolumn{1}{c}{\textbf{Trainable}} & \multicolumn{1}{c}{\textbf{Total}}
& \multicolumn{1}{c}{\textbf{Trainable}} & \multicolumn{1}{c}{\textbf{Total}}
& \multicolumn{1}{c}{\textbf{Trainable}} & \multicolumn{1}{c}{\textbf{Total}} \\
\midrule
VLM
& 0.000 & 1.655
& 0.000 & 1.655
& 0.000 & 1.655 \\

Main DiT
& 0.550 & 0.550
& 0.000 & 0.550
& 0.550 & 0.550 \\

Others
& 0.518 & 0.518
& 0.317 & 0.518
& 0.317 & 0.518 \\

\midrule
Aux encoder
& 0.000 & 1.012
& 0.000 & 1.012
& 0.000 & 1.012 \\

Plug-in branch
& 0.062 & 0.062
& 0.586 & 0.593
& 0.571 & 0.578 \\

\midrule
Total model
& 1.130 & 3.798
& 0.903 & 4.329
& 1.439 & 4.314 \\

Trainable ratio
& \multicolumn{2}{c}{29.76\%}
& \multicolumn{2}{c}{20.87\%}
& \multicolumn{2}{c}{33.35\%} \\
\bottomrule
\end{tabular}
\end{table}

\begin{table}[t]
\centering
\small
\caption{\textbf{Parameter breakdown for the encoder comparison in Sec.~4.2.}
All numbers are in billions of parameters (B).}
\label{tab:param_encoder}
\setlength{\tabcolsep}{3.2pt}
\begin{tabular}{
p{2.3cm}
S[table-format=1.3]@{\,$|$\,}S[table-format=1.3]
S[table-format=1.3]@{\,$|$\,}S[table-format=1.3]
S[table-format=1.3]@{\,$|$\,}S[table-format=1.3]
}
\toprule
& \multicolumn{2}{c}{\textbf{PVI@V-JEPA2}}
& \multicolumn{2}{c}{\textbf{PVI@DINOv2}}
& \multicolumn{2}{c}{\textbf{PVI@DINOv2+V-JEPA2}} \\
\cmidrule(lr){2-3} \cmidrule(lr){4-5} \cmidrule(lr){6-7}
\textbf{Quantity}
& \multicolumn{1}{c}{\textbf{Trainable}} & \multicolumn{1}{c}{\textbf{Total}}
& \multicolumn{1}{c}{\textbf{Trainable}} & \multicolumn{1}{c}{\textbf{Total}}
& \multicolumn{1}{c}{\textbf{Trainable}} & \multicolumn{1}{c}{\textbf{Total}} \\
\midrule
VLM
& 0.000 & 1.655
& 0.000 & 1.655
& 0.000 & 1.655 \\

Main DiT
& 0.000 & 0.550
& 0.000 & 0.550
& 0.000 & 0.550 \\

Others
& 0.317 & 0.518
& 0.317 & 0.518
& 0.317 & 0.518 \\

\midrule
Aux encoder
& 0.000 & 1.012
& 0.000 & 1.136
& 0.000 & 2.149 \\

Plug-in branch
& 0.585 & 0.591
& 0.585 & 0.591
& 0.588 & 0.594 \\

\midrule
Total model
& 0.902 & 4.327
& 0.902 & 4.452
& 0.905 & 5.467 \\

Trainable ratio
& \multicolumn{2}{c}{20.84\%}
& \multicolumn{2}{c}{20.26\%}
& \multicolumn{2}{c}{16.55\%} \\
\bottomrule
\end{tabular}
\end{table}

\FloatBarrier
\section{Full Multi-Task Per-Task Results}
\label{sec:supp_multitask}

This section provides the full per-task results corresponding to Sec. 4.4 in the main paper. 
We report success rates over 100 rollouts for each task in the 20-task and 50-task RoboTwin 2.0 mixtures, where both the fine-tuned GR00T N1.5 baseline and GR00T N1.5 + PVI are trained with 50 demonstrations per task from the same pretrained checkpoint.

As summarized in the main paper, PVI improves the average success rate from 61.15\% to 69.15\% on the 20-task mixture and from 61.32\% to 63.56\% on the 50-task mixture. For completeness, we list the full per-task results below. For readability, the 50-task results are split into two consecutive tables.

\begin{table*}[htbp]
\centering
\small
\caption{\textbf{Per-task success rates on the 20-task mixture.} 
All policies are trained with 50 demonstrations per task and evaluated over 100 rollouts per task. 
PVI improves the average success rate from 61.15\% to 69.15\% (+8.00 pp).}
\label{tab:multitask20_full}
\setlength{\tabcolsep}{10pt}
\begin{tabular}{lcc}
\toprule
\textbf{Task} & \textbf{Baseline} & \textbf{PVI} \\
\midrule
\texttt{adjust\_bottle} & 100\% & 99\% \\
\texttt{beat\_block\_hammer} & 67\% & 96\% \\
\texttt{click\_alarmclock} & 88\% & 99\% \\
\texttt{move\_can\_pot} & 54\% & 52\% \\
\texttt{click\_bell} & 99\% & 97\% \\
\texttt{move\_playingcard\_away} & 61\% & 73\% \\
\texttt{pick\_diverse\_bottles} & 18\% & 60\% \\
\texttt{place\_a2b\_left} & 52\% & 56\% \\
\texttt{open\_laptop} & 68\% & 94\% \\
\texttt{press\_stapler} & 79\% & 84\% \\
\texttt{rotate\_qrcode} & 64\% & 63\% \\
\texttt{scan\_object} & 41\% & 58\% \\
\texttt{put\_object\_cabinet} & 53\% & 62\% \\
\texttt{turn\_switch} & 31\% & 60\% \\
\texttt{handover\_block} & 70\% & 75\% \\
\texttt{open\_microwave} & 17\% & 16\% \\
\texttt{stack\_blocks\_two} & 70\% & 63\% \\
\texttt{stack\_bowls\_two} & 95\% & 85\% \\
\texttt{blocks\_ranking\_rgb} & 69\% & 64\% \\
\texttt{blocks\_ranking\_size} & 27\% & 27\% \\
\midrule
\textbf{Average} & \textbf{61.15\%} & \textbf{69.15\% (+8.00 pp)} \\
\bottomrule
\end{tabular}
\end{table*}

\begin{table*}[htbp]
\centering
\small
\caption{\textbf{Per-task success rates on the 50-task mixture (Part I).} 
All policies are trained with 50 demonstrations per task and evaluated over 100 rollouts per task. 
Across the full 50-task mixture, PVI improves the average success rate from 61.32\% to 63.56\% (+2.24 pp), outperforming the fine-tuned GR00T N1.5 baseline on 32 of 50 tasks and tying on 2 tasks.}
\label{tab:multitask50_full_part1}
\setlength{\tabcolsep}{10pt}
\begin{tabular}{lcc}
\toprule
\textbf{Task} & \textbf{Baseline} & \textbf{PVI} \\
\midrule
\texttt{adjust\_bottle} & 99\% & 100\% \\
\texttt{beat\_block\_hammer} & 43\% & 86\% \\
\texttt{click\_alarmclock} & 85\% & 93\% \\
\texttt{move\_can\_pot} & 43\% & 51\% \\
\texttt{click\_bell} & 98\% & 98\% \\
\texttt{move\_playingcard\_away} & 64\% & 90\% \\
\texttt{pick\_diverse\_bottles} & 48\% & 68\% \\
\texttt{place\_a2b\_left} & 59\% & 52\% \\
\texttt{open\_laptop} & 69\% & 94\% \\
\texttt{press\_stapler} & 92\% & 89\% \\
\texttt{rotate\_qrcode} & 56\% & 63\% \\
\texttt{scan\_object} & 57\% & 56\% \\
\texttt{put\_object\_cabinet} & 73\% & 62\% \\
\texttt{turn\_switch} & 32\% & 68\% \\
\texttt{handover\_block} & 56\% & 67\% \\
\texttt{open\_microwave} & 19\% & 20\% \\
\texttt{stack\_blocks\_two} & 75\% & 48\% \\
\texttt{stack\_bowls\_two} & 84\% & 93\% \\
\texttt{blocks\_ranking\_rgb} & 81\% & 55\% \\
\texttt{blocks\_ranking\_size} & 38\% & 19\% \\
\texttt{dump\_bin\_bigbin} & 88\% & 91\% \\
\texttt{grab\_roller} & 100\% & 98\% \\
\texttt{handover\_mic} & 97\% & 97\% \\
\texttt{hanging\_mug} & 14\% & 19\% \\
\texttt{lift\_pot} & 63\% & 84\% \\
\bottomrule
\end{tabular}
\end{table*}

\begin{table*}[htbp]
\centering
\small
\caption{\textbf{Per-task success rates on the 50-task mixture (Part II).}}
\label{tab:multitask50_full_part2}
\setlength{\tabcolsep}{10pt}
\begin{tabular}{lcc}
\toprule
\textbf{Task} & \textbf{Baseline} & \textbf{PVI} \\
\midrule
\texttt{move\_stapler\_pad} & 6\% & 8\% \\
\texttt{move\_pillbottle\_pad} & 39\% & 63\% \\
\texttt{pick\_dual\_bottles} & 46\% & 62\% \\
\texttt{place\_a2b\_right} & 56\% & 38\% \\
\texttt{place\_bread\_basket} & 67\% & 61\% \\
\texttt{place\_bread\_skillet} & 53\% & 56\% \\
\texttt{place\_burger\_fries} & 95\% & 59\% \\
\texttt{place\_can\_basket} & 52\% & 57\% \\
\texttt{place\_cans\_plasticbox} & 70\% & 71\% \\
\texttt{place\_container\_plate} & 91\% & 96\% \\
\texttt{place\_dual\_shoes} & 38\% & 46\% \\
\texttt{place\_empty\_cup} & 79\% & 88\% \\
\texttt{place\_fan} & 29\% & 33\% \\
\texttt{place\_mouse\_pad} & 26\% & 22\% \\
\texttt{place\_object\_basket} & 76\% & 62\% \\
\texttt{place\_object\_scale} & 46\% & 31\% \\
\texttt{place\_object\_stand} & 66\% & 67\% \\
\texttt{place\_phone\_stand} & 69\% & 72\% \\
\texttt{place\_shoe} & 55\% & 68\% \\
\texttt{put\_bottles\_dustbin} & 48\% & 53\% \\
\texttt{shake\_bottle\_horizontally} & 99\% & 100\% \\
\texttt{shake\_bottle} & 96\% & 100\% \\
\texttt{stack\_blocks\_three} & 32\% & 13\% \\
\texttt{stack\_bowls\_three} & 71\% & 51\% \\
\texttt{stamp\_seal} & 28\% & 40\% \\
\midrule
\textbf{Average} & \textbf{61.32\%} & \textbf{63.56\% (+2.24 pp)} \\
\bottomrule
\end{tabular}
\end{table*}

\end{document}